%% file: access.tex
\documentclass{ieeeaccess}
\usepackage{cite}
\usepackage{amsmath,amssymb,amsfonts}
\usepackage{algorithmic}
\usepackage{graphicx}
\usepackage{textcomp}
\def\BibTeX{{\rm B\kern-.05em{\sc i\kern-.025em b}\kern-.08em
    T\kern-.1667em\lower.7ex\hbox{E}\kern-.125emX}}

\input{preamble}

\begin{document}

\history{Date of publication xxxx 00, 0000, date of current version August 7, 2023.}
\doi{10.1109/ACCESS.2023.3309412}

\title{Multi-NeuS: 3D Head Portraits from Single Image with Neural Implicit Functions}
\author{
\uppercase{Egor Burkov}\authorrefmark{1},
\uppercase{Ruslan Rakhimov}\authorrefmark{1}, 
\uppercase{Aleksandr Safin}\authorrefmark{1},
\uppercase{Evgeny Burnaev}\authorrefmark{1,2},
\uppercase{Victor Lempitsky}\authorrefmark{3}
}

\address[1]{Skolkovo Institute of Science and Technology, 121205 Moscow, Russia}
\address[2]{Artificial Intelligence Research Institute (AIRI), 121165 Moscow, Russia}
\address[3]{No affiliation}
\tfootnote{The work was supported by the Analytical center under the RF Government (subsidy agreement 000000D730321P5Q0002, Grant No. 70-2021-00145 02.11.2021).}

\markboth
{Burkov \headeretal: Multi-NeuS: 3D Head Portraits from Single Image with Neural Implicit Functions}
{Burkov \headeretal: Multi-NeuS: 3D Head Portraits from Single Image with Neural Implicit Functions}

\corresp{Corresponding author: Egor Burkov (e-mail: shrubb@ya.ru).}

\input{sec/0_abstract}
\titlepgskip=-21pt
\maketitle
\input{sec/1_intro}
\input{sec/2_related}
\input{sec/3_method}
\input{sec/4_experiments}
\input{sec/5_conclusion}
\bibliographystyle{unsrt}
\bibliography{refs}
\input{sec/6_authors}
\EOD

\end{document}

%% file: preamble.tex
\usepackage{times}
\usepackage{epsfig}
\usepackage{booktabs}
\usepackage{placeins}

\newcommand{\fig}[1]{Figure~\ref{fig:#1}}
\newcommand{\sect}[1]{Section~\ref{sect:#1}}
\newcommand{\tab}[1]{Table~\ref{tab:#1}}

\usepackage{verbatimbox}

\renewcommand{\paragraph}[1]{\vspace{1mm}\noindent\textbf{#1}}

\usepackage{enumitem}
\setlist[itemize]{noitemsep,leftmargin=*,topsep=0in}
\setlist[enumerate]{noitemsep,leftmargin=*,topsep=0in}

%% file: sec/0_abstract.tex
\begin{abstract}
We present an approach for the reconstruction of textured 3D meshes of human heads from one or few views. Since such few-shot reconstruction is underconstrained, it requires prior knowledge which is hard to impose on traditional 3D reconstruction algorithms. In this work, we rely on the recently introduced 3D representation — neural implicit functions — which, being based on neural networks, allows to naturally learn priors about human heads from data, and is directly convertible to textured mesh. Namely, we extend NeuS, a state-of-the-art neural implicit function formulation, to represent multiple objects of a class (human heads in our case) simultaneously. The underlying neural net architecture is designed to learn the commonalities among these objects and to generalize to unseen ones. Our model is trained on just a hundred smartphone videos and does not require any scanned 3D data. Afterwards, the model can fit novel heads in the few-shot or one-shot modes with good results.
\end{abstract}


\begin{keywords}
3D portraits, 3D reconstruction, few-shot, head reconstruction, meta-learning, neural implicit functions
\end{keywords}

%% file: sec/1_intro.tex
\begin{figure*}[t]
    \centering
    \includegraphics[width=0.99\textwidth]{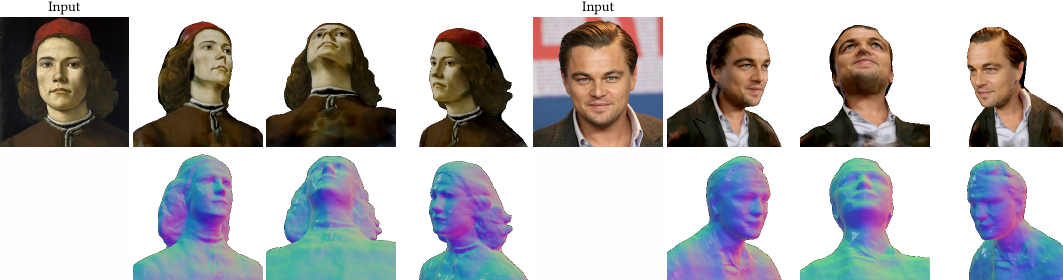}
    \caption{Multi-NeuS can reconstruct a realistic textured 3D mesh of a head from a single in-the-wild photo or painting despite significant domain gap with our training data and the small number of identities in the training set.}
    \label{fig:Teaser}
\end{figure*}

\section{Introduction}
\label{sect:intro}

We consider the task of 3D portraiture, i.e.\ automatic acquisition of 3D models of human heads that capture both the geometry and the texture. This automation avoids costly and time-consuming processes of manual creation of such models. While there is a number of approaches to modeling 2D head appearance \cite{Karras2019stylegan2, deng2019arcface, saponara2021reconstruct}, here we consider 3D head modeling as an important task that finds applications in filmmaking, AR, VR, XR, gaming industries. While a number of learning-based methods for this task have been suggested \cite{H3D-Net,NeuralHeadAvatars}, most of these methods require 3D scans or synthetic data for learning. Here, we propose an alternative approach that learns to model human head shape and appearance directly from a collection of RGB videos.

Our approach is based on a recent class of methods that use implicit representations for shape and appearance such as the recently introduced NeuS method~\cite{NeuS} and very related approaches introduced in parallel with NeuS~\cite{UNISURF,VolSDF,NLR}. We introduce new and simple way to fit such models to individual videos, while sharing a subset parameters, resulting in the approach that we call \textit{Multi-NeuS}. 

We show that sharing the parameters across training videos facilitates knowledge transfer to new individuals unseen during training. As a result, Multi-NeuS achieves noteworthy data-efficiency (capable of learning a generic human head model from the videos of as little as 103 individuals) and is fast to train (takes only 24 hours on a single V100 GPU). After training, Multi-NeuS can create convincing textured 3D head meshes from as little as a single photograph (\fig{Teaser}).


Within this work, we investigate two parameter sharing patterns and sharing-related regularizations that can be used within Multi-NeuS. These are (1) sharing the parameters of layer subsets, as well as more sophisticated (2) low-rank regularization on non-shared parameters. We assess the effect of the sharing setting on the quality of the results.

Overall, in the experiments we show that our system is capable of creating high-quality 3D portraits from few photographs, and reasonably good portraits from single in-the-wild photographs. More generally, our approach proposes a new way to do transfer learning within implicit shape and appearance modeling frameworks, and we hope that our findings will boost future meta-learning research involving implicit functions. To sum up, our contributions are:

\begin{itemize}
\item We introduce a new type of 3D neural implicit architecture that can efficiently fit to many objects of the same class simultaneously and recover their surfaces, given sets of multi-view photos.
\item We devise a meta-learning pipeline for the above model that enables it to reconstruct the textured 3D surface of an unseen object from one or few images.
\item We demonstrate that our system can be applied to single-view reconstruction of 3D full head portraits, producing convincing 3D meshes from in-the-wild images after being trained on just a hundred short smartphone videos.
\end{itemize}

%% file: sec/2_related.tex
\section{Related Work}
\label{sect:related}

\subsection{Neural Implicit 3D Reconstruction}
Neural implicit functions have attracted a lot of attention recently, notably as a flexible approach to represent 3D scenes with neural networks. Contrary to traditional explicit 3D representations such as meshes, they are not limited to a fixed resolution or topology, and, most importantly to us, can naturally employ the power of modern neural network methods.

Neural radiance fields (NeRF) \cite{NeRF} and its extensions (e.g.~\cite{NSVF}) model density and emitted radiance with neural nets which are trained via backpropagating through volumetric ray casting. Although NeRF achieves impressive results in novel view synthesis, it is not designed for reconstructing geometry: meshes directly obtained from NeRF density functions are often full of artifacts \cite{UNISURF}.

A more ``geometry-friendly'' implicit approach is to model the object surface as a zero-level set of implicitly defined occupancy \cite{OccNets} or signed distance function (SDF) \cite{DeepSDF,PatchNets}, which goes back to the classical works on level set reconstruction~\cite{Faugeras98}. The isosurface can then be easily converted to a mesh via marching cubes \cite{MarchingCubes}. To train such models without 3D supervision, several authors have done inverse rendering by modeling color or radiance similar to NeRF and then applying some kind of ray marching \cite{Liu2019,DVR,UNISURF,NLR,VolSDF,NeuS}. From these single-scene multi-view methods, we pick NeuS \cite{NeuS} as a base of our multi-scene few-view method due to simplicity and code availability. We revisit NeuS in more detail in \sect{neus}.

\subsection{Meta-Learning Neural Implicits}
The meta-learning paradigm addresses (among other things) the few-shot problem when given several training examples the network aims to achieve better performance. The most common line of approaches is the optimization-based approaches \cite{finn2017model,nichol2018first} that learn the best weight initialization. For a deeper meta-learning review we refer the reader to \cite{hospedales2020meta}. Regarding the application of meta-learning to neural implicits, MetaSDF \cite{sitzmann2020metasdf} exploits this idea to learn the initialization of the SDF network, while the work~\cite{tancik2021learned} applies meta-learning to a wider variety of signal types. Our work concentrates on human body representation and uses shared network layers across different tasks (with different people identities).

\subsection{Few- or Single-View Head Reconstruction}
Historically, directly fitting statistical 3D Morphable Models (3DMMs) to an image has been a popular method to recover the 3D head shape \cite{PiotraschkeBlanz2016,JiangLiu2017,HuHuber2017,GANFIT}, but 3DMMs are limited to coarse shape estimation, requiring separate steps of reconstructing e.g.~wrinkles \cite{JiangLiu2017} or hair \cite{BaoZhang2020}.
In addition, 3DMMs are constructed from 3D scans which might be hard to obtain for many classes.
Other more descriptive and flexible 3D representations include depth maps \cite{Pix2Vertex,Deep3DPortrait,LiftedGAN}, regular meshes \cite{Szabo2019,Rotger2019}, and volumetric grids \cite{VRN}, although many of these approaches still rely on 3DMM in their intermediate steps.
Two rare examples of completely model-free methods that also reconstruct hair \cite{LiftedGAN,Szabo2019} are self-supervised GANs \cite{StyleGAN2} that learn from unlabeled collections of images.
However, actual fitting to unseen images (e.g.~GAN inversion) was not demonstrated.
More information on face/head reconstruction before the advent of 3D neural implicit methods is available in recent comprehensive surveys \cite{Morales2020,Egger2020}.

Recently, several works have successfully applied neural implicit representations to the head reconstruction task, but most of them either do not reconstruct geometry directly (e.g.~because of ill-suited NeRF representation) or require complex datasets.
Portrait-NeRF \cite{Portrait-NeRF} is an early attempt of meta-learning a single-view NeRF. The support of only slight viewpoint changes has been demonstrated for this method.
The i3DMM method~\cite{i3DMM} introduced the first 3DMM to include hair. This method is based on SDFs and is constructed from about 2000 3D scans of 64 people.
H3D-Net \cite{H3D-Net} meta-learns high-quality SDF representation of full static heads and supports reconstruction from as few as three posed images (though feeding just one image is also possible). The method is trained on a private dataset of 10,000 structured-light 3D scans.
HeadNeRF \cite{HeadNeRF} yields controllable NeRF portraits conditioned on latent 3DMM vectors (identity, expression, albedo, and illumination). It is a fully supervised approach, and authors were able to train it in reasonable time thanks to a strategy that improves rendering performance~\cite{StyleNeRF}.
The authors of EG3D \cite{EG3D} went even further and trained a StyleGAN2 \cite{StyleGAN2} to yield volume-renderable 3D heads with very little supervision (a similar idea was proposed in VolumeGAN \cite{VolumeGAN} simultaneously). Like HeadNeRF, EG3D can fit an arbitrary head photo by optimizing the latent vector(s). Moreover, their paper demonstrates extracting meshes of convincing quality. Still, this method is computationally $80 \times$ more expensive to train than Multi-NeuS, and it does not reconstruct parts of the head that are further from the face due to the lack of dedicated background modeling.

%% file: sec/3_method.tex
\begin{figure*}[!hbt]
    \centering
    \includegraphics[trim={231 320 140 32},clip,width=0.95\textwidth]{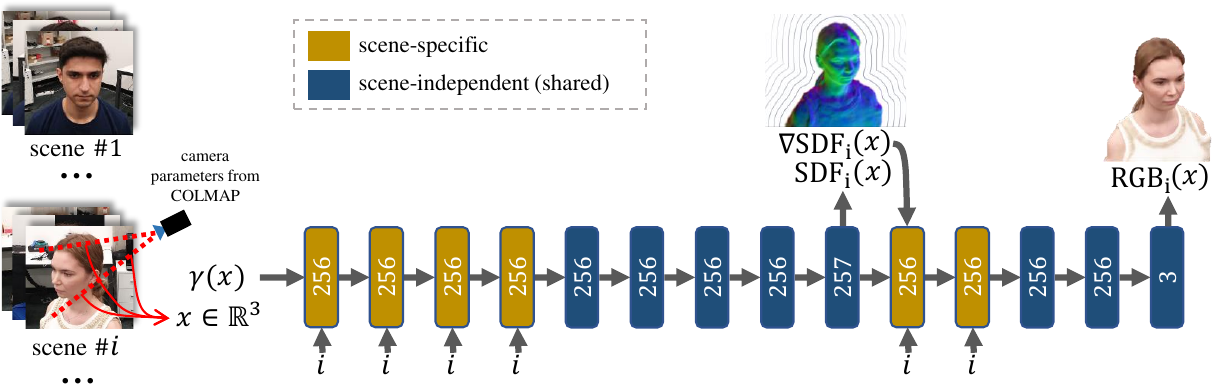} 
    \caption{Architecture of Multi-NeuS, a 3D neural implicit function that can represent multiple objects of a class simultaneously (boxes depict fully connected layers and their output dimensionalities; $\gamma$ is the positional encoding function). Since some layers (blue) are shared between all scenes, they can learn class priors to then transfer knowledge to novel scenes of the same class, enabling few-shot reconstruction. The model is trained via volumetric rendering and simple pixelwise loss, just like NeuS \cite{NeuS}, but on a dataset of multiple scenes. Afterwards, when fitting to an unseen object, scene-specific layers (yellow, \sect{layers}) are fitted first, and finally all layers are fine-tuned together.}
    \label{fig:pipeline}
\end{figure*}

\section{Method}
\label{sect:method}

\subsection{Recap: NeuS Reconstruction}
\label{sect:neus}

As our method builds upon NeuS~\cite{NeuS}, we start with the review of this method. NeuS is a modification of NeRF \cite{NeRF} for non-transparent objects. It models the object surface directly, thus allowing 3D surface reconstruction from images using differentiable neural rendering. Specifically, the object surface in NeuS is represented as the zero-level set of a signed distance function $\left\{ x \in \mathbb{R}^3 ~|~ \text{SDF}(x) = 0 \right\}$, where SDF is defined as signed distance to object surface and is modeled by a neural network. In addition, RGB \textit{radiance} at any 3D point is modeled by another neural net, and \textit{density} is modeled as a bell-shaped function of SDF that attains its maximum at zero, i.e.\ at the object surface. More specifically (see \fig{pipeline}), the SDF network is a simple multi-layer perceptron (MLP) with 8 hidden layers of 256 neurons and softplus  activations ($\beta=100$), and the radiance network is an MLP with 4 hidden layers of 256 neurons and ReLU activations. The former network takes a 3D coordinate and outputs an SDF value and a latent vector. Meanwhile, the latter network takes this latent vector, the 3D coordinate, the camera view direction, and the gradient of the SDF, and outputs the RGB radiance value. Positional encodings~\cite{NeRF} are applied to 3D coordinates (6 dimensions) and view directions (4 dimensions).

The radiance and density of points sampled along the rays corresponding to pixels of input images are used to run differentiable volume rendering \cite{NeRF} that integrates the samples along the ray and outputs its RGB color.
The optimization algorithm forces the RGB results of ray integration to be similar to the corresponding known pixel intensities by progressively tuning the weights of neural networks. The loss function to optimize is a simple pixelwise mean squared error combined with an eikonal regularization term that ensures $\left\| \nabla\text{SDF}(x) \right\| = 1$. After convergence, it is possible to obtain object mesh via marching cubes \cite{MarchingCubes} over $\text{SDF}(x)$, as well as to synthesize novel views by volume rendering or any ray marching algorithm, such as sphere tracing.

The multi-view captures may include distant background which is difficult to represent by the above neural nets. Therefore, the object of interest is considered to be within a unit sphere, and everything outside of that sphere is modeled by a separate dedicated NeRF with the special parametrization of coordinates \cite{NeRF++}. To optimize this NeRF along with NeuS, extra ray points are sampled outside of the unit sphere. A sufficiently large dataset lets such tandem to disentangle background from the central object automatically, without mask supervision.

NeuS achieves excellent results when applied to sets containing dozens of images. Our goal is to create a NeuS-based system that can perform reconstruction given a single image or very few images. This scenario is too under-constrained for the original NeuS and will result in poor convergence. To alleviate this, we narrow down the class of potential scenes to human heads and pre-train our model on a dataset of multiple people, while facilitating knowledge transfer to unseen people as discussed below.

\subsection{Multi-NeuS}

Our solution called Multi-NeuS is depicted in \fig{pipeline}. We upgrade NeuS so that it can fit to $N$ scenes simultaneously. Our high-level idea is simple. We create $N$ copies of scene-specific NeuS instances that share \textit{some} of the layers, while keeping other layers unshared (\textit{scene-specific}). We then fit these $N$ instances to the scenes simultaneously, while optionally imposing additional structural regularization on scene-specific layers. 

Naturally, we expect that during such fitting shared layers will tend to model features useful to represent any object, while scene-specific layers combine, refine and augment the output of shared layers to model a specific object. For instance, a shared layer might model rough basic human head shapes, while the following (scene-specific) layer may learn the weights with which to combine those shapes, like in linear blend skinning models.

We experiment with two architectures for scene-specific layers that are described below in \sect{layers}. As shown in \fig{pipeline}, we use scene-specific layers in the first halves of the SDF network and the radiance network, while sharing all other layers. This choice is evaluated in \sect{Layers position}.

Differently from NeuS, Multi-NeuS learns $N$ independent scene-specific instances of background NeRFs. Also, we do not model view-dependent effects in our architecture, effectively assuming that human heads do not produce specular reflections. We find that on our dataset (which is captured in scattered light), this does not hurt validation performance but significantly reduces overfitting in few-shot mode especially when generalizing to new lighting.

\subsection{Scene-Specific Layers}
\label{sect:layers}

\begin{figure}[!ht]
    \centering
    \includegraphics[trim={220 82 437 294},clip,width=0.47\textwidth]{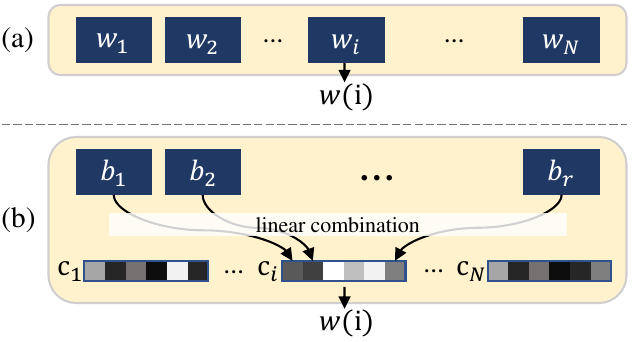}
    \caption{The two architectures of scene-specific layers explored in our paper, \textit{independent} (a) and \textit{low-rank} (b). They are fully connected layers whose weights and biases $w\left(i\right)$ depend on scene index $i$. An \textit{independent} layer learns individual weights and biases for each of $N$ scenes, while a \textit{low-rank} layer learns $r$ copies and then linearly combines them with each scene's own learnable coefficients.}
    \label{fig:Scene-specific layers}
\end{figure}

We use the scene (person) index $i \in \overline{1,N}$ to enumerate scene-specific layer instances. Thus, by considering different instances within scene-specific layers, the same network architecture models every object in the dataset. In this work, we experiment with two architecture choices for scene-specific layers (\fig{Scene-specific layers}), which we term \textit{independent} and \textit{low-rank}. They are described below.

\paragraph{Independent layers} (\fig{Scene-specific layers}, a). This is a straightforward implementation where the scene-specific layer has a dedicated set of weights and biases $w_1, \dots, w_N$ for each scene. This architecture has large representational power but has significant drawbacks.

First, during meta-learning, each $w_i$ receives infrequent weight updates during learning. Thus, if a training minibatch includes pixels from few ($m \ll N$) scenes, then sub-layers corresponding to all other scenes do not receive any weight updates. Alternatively, a minibatch can be composed of random pixels from the entire dataset ($m \approx N$). In this case, however, $w_i$'s gradients become too noisy, coming from just few ($\approx \frac{N}{m}$) pixels, again leading to slow/poor convergence.

In practice, batching together pixels from many scenes is inefficient as it requires to run $m \approx N$ layers in each forward pass, so in our experiments we set $m = 1$, i.e.\ we sample all pixels of a minibatch from just one scene. We therefore use the Adam \cite{Adam} optimizer but update moment statistics for a scene-specific layer only when the corresponding scene participates in the forward pass (known as ``sparse/lazy Adam").

Another related problem with \textit{independent} layers is overfitting due to the excessive number of parameters. This often leads to poor generalization to new subjects. Our second architecture below is designed to alleviate this by a built-in regularization.

\paragraph{Low-rank layers} (\fig{Scene-specific layers}, b). In this scheme, scene-specific layer's weights and biases $w\left(i\right) \in \mathbb{R}^p$ are not learnt directly. Instead, they are computed as a linear combination of $r$ \textit{basis vectors} $b_1, \dots, b_r$:

\begin{equation}
\label{eq:Low-Rank Layer}
w\left(i\right) = \sum_{j=1}^r c_{ij} {b_j} ~,
\end{equation}

where $r$ is the layer's \textit{rank}. We learn both the basis vectors $b_j \in \mathbb{R}^p$ and the linear combination coefficients $c_{ij} \in \mathbb{R}$, where $i \in \overline{1,N}$ and $j \in \overline{1,r}$. Thus, each scene-specific layer learns a single set of $r$ basis vectors for the entire training dataset containing multiple scenes, and these vectors are recombined with different weights to model different scenes, therefore a separate set of $r$ coefficients is learned for each of the $N$ scenes. Such low-rank factorization decreases the number of parameters significantly (by a factor of several hundreds in our experiments), reducing overfitting. 





\subsection{Training}
\label{sect:Training}

\begin{figure*}[!ht]
    \centering
    \includegraphics[trim={10 220 180 10},clip,width=0.82\textwidth]{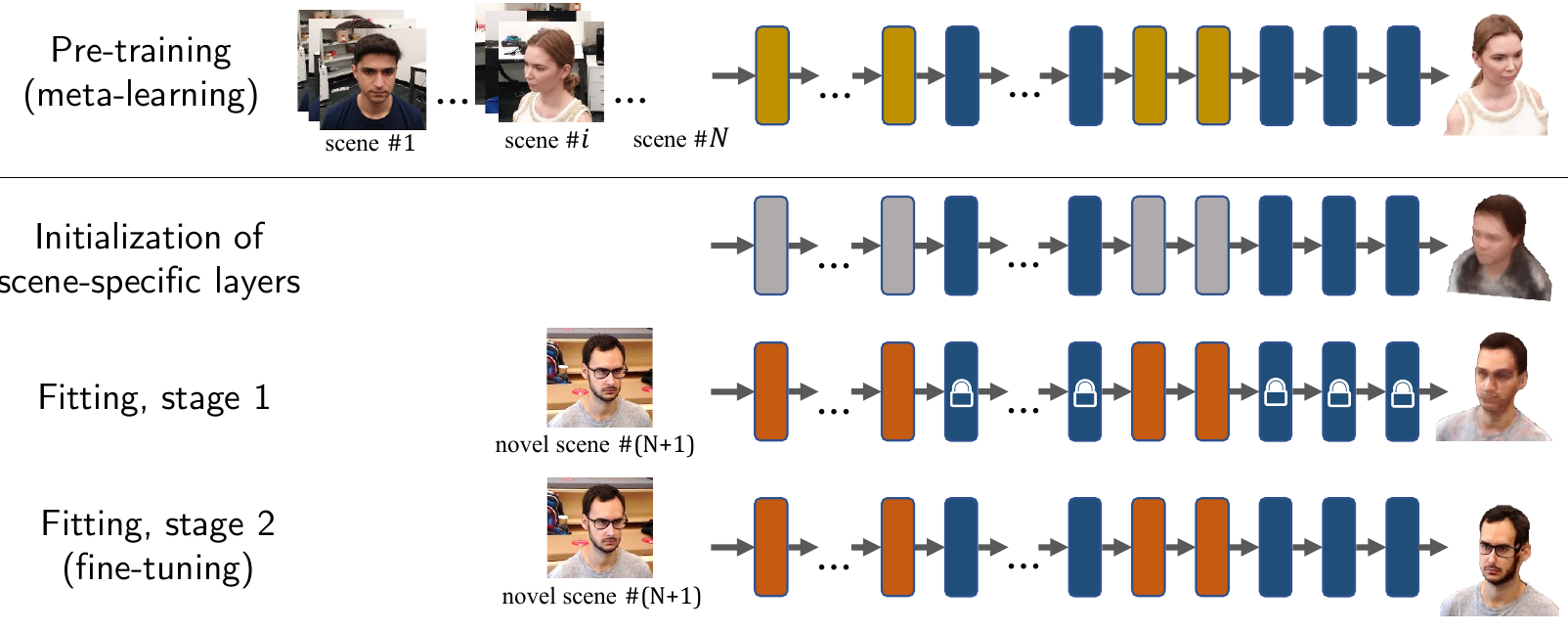} 
    \caption{The training stages of Multi-NeuS (\sect{Training}). Row 1: the entire model, including scene-specific layers, is optimized to represent $N$ scenes. Row 2: to fit to a novel ("$(N+1)$-st") scene, scene-specific layers are re-initialized by the aggregation over the weight values over $N$ scenes. Row 3: scene-specific layers only are optimized for the novel person (based on as few as one or two images). Row 4: all layers are optimized with a smaller learning rate.}
    \label{fig:Training diagram}
\end{figure*}

Multi-NeuS is applied in two stages: meta-learning and fitting (see \fig{Training diagram}).

In the initial \textbf{meta-learning} stage, we pre-train the whole architecture using the same volumetric rendering procedure as in NeuS (\sect{neus}) but on a dataset of multi-view RGB images of $N$ scenes rather than a single scene. At every optimization step, a minibatch of camera rays (or, equivalently, image pixels) is sampled uniformly from eight random images of one random scene. Eventually, Multi-NeuS estimates the 3D shape and texture of every scene (subject) in the dataset. 

After meta-learning, we can fit to new scenes starting from the pre-trained initialization. This \textbf{fitting} stage is thus conducted to estimate the 3D shape and the texture of a novel unseen object. To represent that object, we add the new $(N+1)$-st scene to the model, that is, the $(N+1)$-st set of scene-specific layers, initialized as described below. This time, we are given images of the new subject (can be as few as one or two), their estimated camera parameters, and their background segmentation masks.

The fitting process is performed in two steps: we first retrain the scene-specific weights and then we fine-tune all weights to the new scene. The first step begins with initializing scene-specific weights. We do it by simply averaging these weights over $N$ scenes so that the $(N+1)$-st representation in Multi-NeuS essentially represents ``the average object in the dataset'' as learned by the scene-specific layers. That is, for \textit{independent} layer we set $w_{N+1} = \frac{1}{N} \sum_{i=1}^N w_i$, and for the \textit{low-rank} layer $c_{N+1} = \frac{1}{N} \sum_{i=1}^N c_i$. Note that in the \textit{low-rank} layer the basis weights $b_1, \dots, b_r$ are not scene-specific but are in fact shared by all scenes. Therefore, we do not optimize or reset them in the first step of the fitting process. After optimizing the newly initialized scene-specific weights in the first step, in the second step we ``unfreeze'' the shared weights and optimize all weights while using a smaller learning rate.

The optimization during the fitting stage is performed in the same way as in the meta-learning stage with two notable differences. Firstly, instead of using a dedicated background NeRF~\cite{NeuS}, we explicitly estimate background masks and optimize an additional loss that forces the SDF isosurface to match these masks. The loss used in this case is the binary cross-entropy between the accumulated density over a ray and the foreground mask value ($1$ if object, $0$ if background). This is needed since we found that background separation in the original NeuS works unsatisfactory in the few-shot regime.

The second modification is fine-tuning the camera parameters. This is needed because camera estimates can be inaccurate, especially for in-the-wild images from the Internet. To compensate for that, we backpropagate the losses into the camera parameters and optimize them alongside the neural networks with a $10\times$ smaller learning rate.

Please refer to \sect{Training details} for additional details, including the implementation details and the hyperparameters.

%% file: sec/4_experiments.tex
\begin{figure*}[ht]
    \centering
    \includegraphics[width=0.99\textwidth]{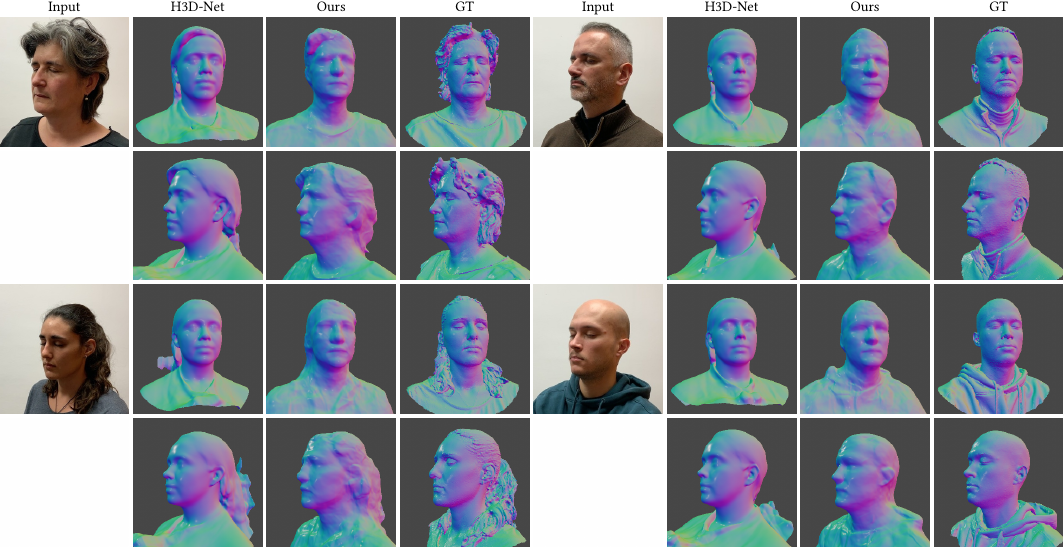}
    \caption{Single-view mesh reconstruction on the first four scenes of the H3DS dataset. H3D-Net \cite{H3D-Net}, a method related to ours, was designed for three-view reconstruction but can also be evaluated in the one-shot mode. The H3D-Net system was trained on 10,000 3D scans from the same distribution as these test examples. Our method is trained on a hundred smartphone videos and still matches the quality of H3D-Net, while demonstrating somewhat smaller identity gap and less pronounced regression-to-mean effect.}
    \label{fig:H3DS, Ours vs H3D-Net}
\end{figure*}

\section{Experiments}
\label{sect:experiments}

\subsection{Datasets}
\label{sect:datasets}

Our \textbf{training} (meta-learning) dataset is a subset of SmartPortraits \cite{SmartPortraits}. It consists of 107 short ($\approx$ 25 seconds) smartphone videos of still people with neutral pose and facial expression. Four of these (two female and two male subjects) serve as the validation set. In each video, the distance to the head ($\approx$ 1.5 m) and the elevation are roughly constant, while the azimuth travels within $\pm 45^{\circ}$. From each video, we remove frames with flash and randomly pick about 77 frames from the rest, shrinking the entire dataset to 8256 images. We obtain camera parameters by running the COLMAP structure-from-motion software \cite{COLMAP} on these images. Finally, these images are loosely cropped to head and shoulders using a face detector. Note that we do not use any motion or depth information in our system.

Because Multi-NeuS takes in the absolute 3D coordinates, all scenes are aligned against each other to minimize the relative difference between objects. This helps our network not to spend capacity on modeling translations and scaling, and thus to fit the training set easier. We accomplish approximate alignment as follows. For each scene, we detect six prominent facial landmarks in images \cite{FaceAlignment}. We then triangulate the 2D landmarks to get their 3D coordinates. We choose the the first scene of SmartPortraits as a reference one. For all other scenes we compute an optimal similarity transform $T$ \cite{Umeyama1991} that aligns two set of points: the triangulated 3D landmarks with the reference ones. It is achieved by finding the optimal translation, rotation and scaling by minimizing the root-mean-square deviation of the point pairs. Finally, the transform $T$ is applied to all camera poses of the current scene. We estimate and apply such similarity transform not only for SmartPortraits but for every scene of every dataset used in this work.

We also \textbf{validate} on the H3DS dataset \cite{H3D-Net}, which consists of ten individuals. For each individual, the dataset offers a full head 3D scan (mesh) alongside with 60 to 70 $360^{\circ}$ photos taken with varying lighting, and camera parameters for these photos.

Besides, we provide qualitative results on several paintings and in-the-wild photos found on the Web. To that end, we estimate camera parameters for a single photo as follows. We detect the same six landmarks as above, but this time obtain their approximate 3D coordinates in orthographic camera coordinate system (\cite{FaceAlignment} provides them directly). We assume that these coordinates are 3D world coordinates, and that the image was taken with a telephoto lens with the vertical field of view of $\approx 10^{\circ}$. These asssumptions allow us to roughly recover the camera pose in world coordinates, namely via an algorithm for the Perspective-n-Point (PnP) problem \cite{PNP_IPPE}.

When fitting to any unseen pictures, we estimate background masks using an off-the-shelf model \cite{Graphonomy} and manually refine them.

\subsection{Single-View Geometry Reconstruction}
\label{sect:Single-View Reconstruction}

By providing ground truth 3D scans, H3DS permits a quantitative comparison of geometry reconstruction, so we use it to compare against H3D-Net \cite{H3D-Net}, which was tailored for this dataset. Although H3D-Net was demonstrated to reconstruct from three or more views, it can fit to a single view as well, and it has an advantage on the H3DS dataset since this model was trained on a large dataset (10,000 scenes) from the same distribution. 

The target metrics, as in \cite{H3D-Net}, are unidirectional Chamfer distances in millimeters from the predicted mesh to the ground truth, computed after rigid alignment via ICP \cite{ICP}. One metric is the distance computed over facial area only, and the other one is computed over the entire ground truth mesh of a head.

We compute 1-view metrics by reconstructing from left, right (azimuth $\approx 45^{\circ}$), and frontal views. We do not apply our method in few-shot setting on H3DS because images in this dataset are taken with varying lighting and exposure, lacking multi-view consistency required for Multi-NeuS.

We compare our best model (low-rank architecture, $r = 1000$; evaluated in \sect{Effect of Number of Views and Layer Type}) with H3D-Net in \tab{Ours vs H3D-Net} and \fig{H3DS, Ours vs H3D-Net}. Our method practically matches H3D-Net in reconstruction accuracy while learning from a different dataset that has $100\times$ fewer identities and does not require 3D scanning. Furthermore, rendered samples suggest that H3D samples look very similar to each other, especially outside of the face region (the so-called regression-to-mean effect) while our model predicts more ``personalized'' shapes.

\setlength\tabcolsep{1.5pt} 
\begin{table}[t]
  \begin{center}
    {\small{
\begin{tabular}{l|cccc|cccc|}
\multicolumn{1}{c}{} & \multicolumn{4}{c}{face} & \multicolumn{4}{c}{head} \\
\cline{2-9}
\textit{Input view} & \textit{F} & \textit{L} & \textit{R} & \textit{mean} & \textit{F} & \textit{L} & \textit{R} & \textit{mean} \\
\toprule
H3D-Net 3-view & - & - & - & 1.34 & - & - & - & 10.53 \\
\midrule
H3D-Net 1-view & \textbf{1.82} & 1.83 & 1.91 & 1.85 & 13.83 & \textbf{13.01} & 12.51 & 13.12 \\
Ours 1-view & 1.89 & \textbf{1.77} & \textbf{1.86} & \textbf{1.84} & \textbf{13.00} & 13.27 & \textbf{11.95} & \textbf{12.74} \\
\bottomrule
\end{tabular}
}}
\end{center}
\caption{Mesh reconstruction error in millimeters on H3DS dataset. Lower is better, "F/L/R" are for "frontal/left/right". See \sect{Single-View Reconstruction} for details.}

\label{tab:Ours vs H3D-Net}
\end{table}

To demonstrate additional single-view geometry reconstruction, we show several reconstructions of in-the-wild photographs and paintings in \fig{In-the-wild reconstruction} and in the supplementary video. 

\subsection{Effect of Number of Views and Layer Type}
\label{sect:Effect of Number of Views and Layer Type}

\begin{figure}[!t]
    \centering
    \includegraphics[trim={40 140 193 25},clip,width=0.47\textwidth]{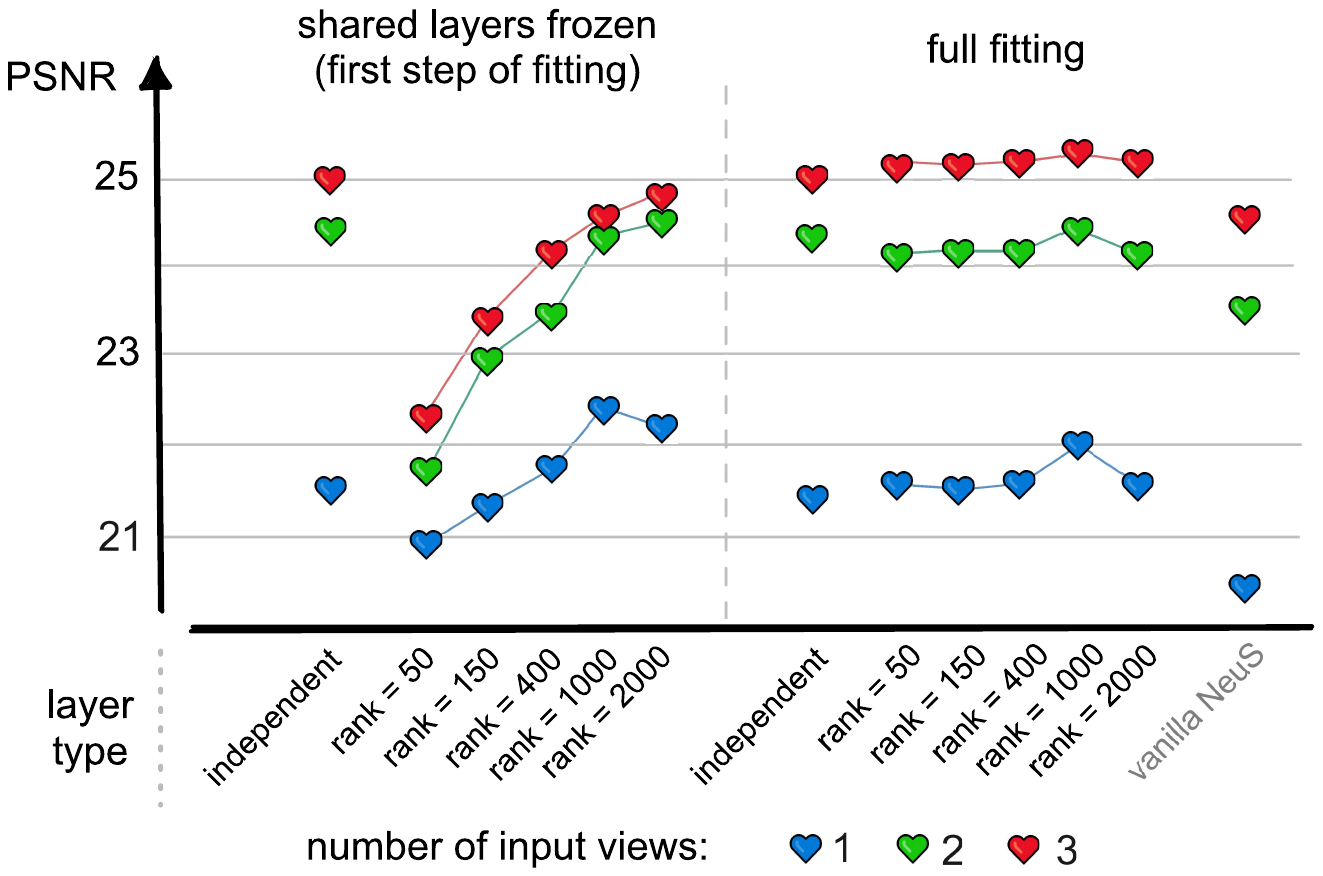}
    \caption{Quality of novel view reconstruction depending on scene-specific layer type (\sect{layers}), measured on our validation subset of SmartPortraits. Lower rank metamodels have fewer degrees of freedom and underfit during the first step of fitting; higher rank models fit better and provide a more convenient initialization for the second step of fitting (fine-tuning of all weights).}
    \label{fig:Number of views vs layer type}
\end{figure}

Although our primary aim is to reconstruct heads given just one image, our method naturally benefits from additional views. We demonstrate this on the validation subset of SmartPortraits. Similarly to H3DS, we restrict the scenes to the views: left, right (with azimuths around $\pm 45^{\circ}$) and frontal. Since 3D ground truth is not available in this case, we render two additional \textit{control} views ($\pm 20^{\circ}$) and compute masked PSNR against the ground truth images corresponding to these two views (\textit{control images}). These are in turn averaged over four validation scenes.

\begin{figure*}[ht]
    \centering
    \includegraphics[width=0.99\textwidth]{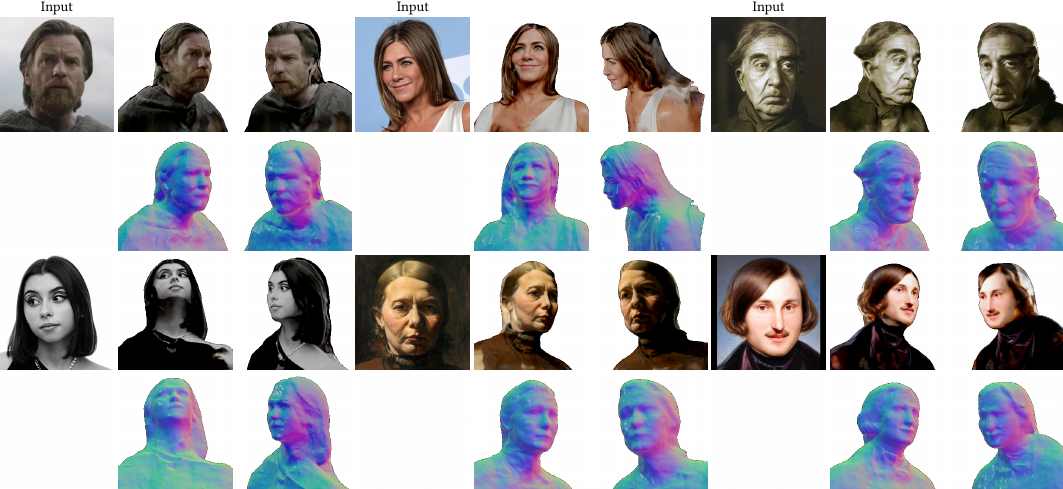}
    \caption{Additional results for 3D reconstruction of in-the-wild photographs and paintings. Our method is able to handle different hair styles and performs reasonably well for images that are different from the training SmartPortraits dataset. Note the back may have artifacts because this part of the head is mostly absent in the training dataset due to very limited view angles.}
    \label{fig:In-the-wild reconstruction}
\end{figure*}

We observe that during fitting to a novel person, optimizing camera parameters provides additional degrees of freedom. This often leads to the person's shape in Multi-NeuS drifting away from its ``canonical'' position (\sect{datasets}) and inflating the validation error, even when the reconstruction is good. Moreover, the two control cameras might be estimated inaccurately during data pre-processing. To address this, before reporting PSNR against control images, we refine the control cameras' poses and focal distances by optimizing for PSNR.

\fig{Number of views vs layer type} compares how faithfully the novel views are reconstructed depending on the number of input views (one, two, or three) and depending on the layer type (\textit{independent} or \textit{low-rank}). In the case of a single input view, the metric is averaged over reconstructions from left, right, and frontal views. In the two views case, we take the left+right views as input.

\begin{figure*}[!ht]
    \centering
    \includegraphics[trim={14 1435 14 1085},clip,width=0.99\textwidth]{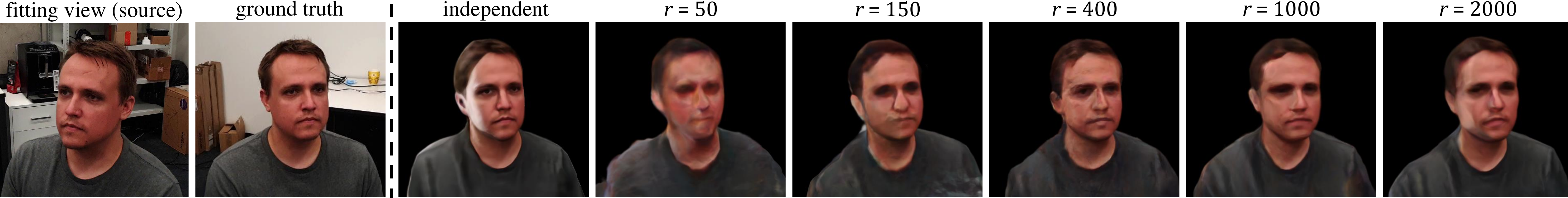}
    \caption{The \textbf{first (coarse) step} of fitting (i.e. shared layers frozen) for various architectures of scene-specific layers. The scene is a subject from the validation subset of SmartPortraits. The amount of overfitting can be traced by looking at the right cheek and ear, which are invisible in the source view. The \textit{independent} architecture with overparametrized scene-specific layers overfits already at this fitting stage. The \textit{low-rank} variants become better at fitting these hidden parts with increasing ranks and model the texture better, but at some point ($r = 2000$ in this case) get too many degrees of freedom and start to overfit. $r$ = 1000 provides optimal reconstruction in this case (and on average).}
    \label{fig:SmartPortraits}
\end{figure*}

Consider \textit{low-rank} models. Clearly, at the first step of fitting (\fig{Number of views vs layer type}, left), when only linear combination coefficients $c_{N+1}$ and camera parameters are optimized, the models underfit in the case of low ranks. When the rank is very high (2000), the models start to overfit since the number of parameters becomes excessive. This is additionally illustrated in \fig{SmartPortraits}.

In all cases, the second step of fitting (\fig{Number of views vs layer type}, right) where all parameters are fine-tuned is necessary because low-rank coefficients have few degrees of freedom. Multi-NeuS without the second step thus underfits the input views. However, in the few-shot setting, optimizing the full network can lead to severe overfitting. So the primary goal of the first fitting step is to provide a good initialization for this second step. According to the diagram, models with a reasonably high rank (e.g.~1000) provide best initializations, but this comes at a cost of overfitting during fine-tuning which may even decrease the overall PSNR by distorting the unseen areas of the resulting shape. This is probably because high rank scene-specific layers gain too much representational power and the shared layers are not forced as hard to learn universal features, thus hampering generalization. An alternative interpretation is that since we spend the same number of fine-tuning iterations regardless of rank, early stopping might alleviate some overfitting issues.

Another obvious observation is that with more views the effect of overfitting decreases, and the advantage of higher-rank models becomes less pronounced (e.g.~a 50-rank model already does well for the three-view reconstruction). In addition, the change in the number of views allows to assess the capacity of scene-specific layers.

The model with \textit{independent} scene-specific layers does not generalize well because of the excessive capacity. Although it demonstrates larger PSNR than 50- and 150-rank models, it does so because its scene-specific layers are usual linear layers which can fit the training view really well, while low-rank models underfit. At the same time, the unseen parts in the validation views already get distorted in the first step and this is why the second step (fine-tuning) does not improve the score in this case.

Finally, to prove the necessity of shared architectures and meta-learning, we compare to a simple baseline (\fig{Number of views vs layer type}, extreme right) where a vanilla NeuS (without view directions) is trained on a scene from SmartPortraits and is then fine-tuned (transfer-learned) in a few-shot scenario to the target scene. This is essentially equivalent to Multi-NeuS with $N = 1$, i.e. with a pre-training dataset of 1 scene. The score for this baseline was computed by transferring from 4 different SmartPortraits training scenes (2 male, 2 female) and averaging the metric. Although NeuS typically fits better to a single scene than Multi-NeuS to any of its meta-learning scenes, its few-shot generalization ability is clearly lower compared to any version of Multi-NeuS.

\subsection{Which Layers to Make Scene-Specific}
\label{sect:Layers position}

\setlength\tabcolsep{2pt} 
\begin{table}
  \begin{center}
    {\small{
\begin{tabular}{lccc|ccc}
\multicolumn{1}{c}{} & \multicolumn{3}{c|}{one step of fitting} & \multicolumn{3}{c}{full fitting} \\
\multicolumn{1}{l}{\textit{Number of views}} & \textit{1} & \textit{2} & \textit{3} & \textit{1} & \textit{2} & \textit{3} \\
\toprule
{\scriptsize $\blacksquare \blacksquare \blacksquare \blacksquare \blacksquare \blacksquare \blacksquare \blacksquare \blacksquare ~ \blacksquare \blacksquare \blacksquare \blacksquare \blacksquare$} & \multicolumn{6}{c}{did not converge} \\
{\scriptsize $\square \square \square \square \square \square \square \square \square ~ \blacksquare \blacksquare \blacksquare \blacksquare \blacksquare$} & 17.61 & 19.35 & 20.78 & 18.19 & 19.21 & 23.67 \\
{\scriptsize $\square \blacksquare \square \square \blacksquare \square \square \blacksquare \square ~ \square \blacksquare \square \square \blacksquare$} & 20.49 & 23.10 & 23.71 & 20.76 & 24.03 & 25.25 \\
{\scriptsize $\square \blacksquare \square \blacksquare \square \blacksquare \square \blacksquare \square ~ \square \blacksquare \square \blacksquare \square$} & 21.48 & 23.72 & 24.18 & 21.26 & 24.19 & 25.16 \\
{\scriptsize $\square \square \square \square \square \blacksquare \blacksquare \blacksquare \blacksquare ~ \square \square \square \blacksquare \blacksquare$} & 21.75 & 23.94 & 24.29 & 20.78 & 24.31 & 25.15 \\
{\scriptsize $\blacksquare \blacksquare \blacksquare \blacksquare \square \square \square \square \square ~ \square \square \square \square \square$} & 21.85 & 23.85 & 24.56 & 21.84 & 24.28 & 25.23 \\
{\scriptsize $\blacksquare \blacksquare \blacksquare \blacksquare \blacksquare \blacksquare \blacksquare \blacksquare \blacksquare ~ \square \square \square \square \square$} & 21.92 & 24.25 & \textbf{24.77} & 21.59 & 24.03 & 25.22 \\
{\scriptsize $\blacksquare \blacksquare \blacksquare \blacksquare \square \square \square \square \square ~ \blacksquare \blacksquare \square \square \square$} & \textbf{22.43} & \textbf{24.26} & 24.66 & \textbf{21.99} & \textbf{24.39} & \textbf{25.27} \\
\bottomrule
\end{tabular}
}}
\end{center}
\caption{Novel view reconstruction quality depending on the choice of layers to replace with their scene-specific variants. We test the performance on the validations scenes from SmartPortraits and report PSNR values (in dB) averaged across holdout views. Boxes depict sequential fully-connected layers --- like in NeuS, there are 9 layers that predict SDF, followed by 5 layers that predict radiance. $\blacksquare$ means scene-specific layer, $\square$ means shared, i.e.~vanilla linear layer. Layer type is \textit{low-rank}, $r$ = 1000.}

\label{tab:Layers position}
\end{table}

In this subsection, we evaluate the exact choice to put scene-specific layers into the first halves of the SDF network and the radiance network of vanilla NeuS.

Some possible choices are listed in \tab{Layers position} and are evaluated for \textit{low-rank} shared layers with $r$ = 1000. Putting the scene-specific layers to the radiance network only results in a constant SDF in the first stage of fitting. This results in very low metrics. Results for other sharing patterns are harder to interpret, but arguably good performance requires sufficient number of scene-specific layers (having too few of them is detrimental). Furthermore, at least some of these layers should be among the early processing layers.

\subsection{Implementation Details and Hyperparameters}
\label{sect:Training details}

In both meta-learning and fitting, we use minibatches of 512 rays. In our experiments, there are 610,000 optimization iterations during meta-learning, 12,000 iterations in the first stage of fine-tuning and 13,000 in the second stage of fine-tuning. Pre-training (meta-learning) Multi-NeuS takes around 24 hours and fitting it to a novel subject takes about an hour on a single NVIDIA V100 GPU. The learning rate is $1.8 \cdot 10^{-4}$ in meta-learning, $4 \cdot 10^{-4}$ in the first step of fitting and $6 \cdot 10^{-5}$ in the second step. We multiply the learning rate by $0.316$ every time the loss stops decreasing (known as ``reduce-on-plateau schedule"). All other hyperparameters, including the number of ray sampling steps, eikonal loss weight, weight initialization (including that in the fitting stage) are kept the same as in NeuS \cite{NeuS}.

We optimize camera parameters similar to \cite{lin2021barf}. Specifically, we (1) multiply initial camera rotation matrix by optimizable update matrix parametrized using $\mathfrak{so}(3)$ Lie algebra, (2) add a optimizable residual to the translation parameters, and (3) multiply focal length by an optimizable scalar.

%% file: sec/5_conclusion.tex
\section{Discussion}
\label{sect:conclusion}

We have presented \textit{Multi-NeuS} -- an approach for one- and few-shot 3D head portrait reconstruction. The approach can reconstruct head portraits in the form of surface mesh and texture. To enable the few-shot capability, we propose and validate a very simple idea of taking a scene-specific deep architecture (NeuS) and fitting it to multiple scenes, while sharing some parameters across scenes. We show that despite simplicity, this idea is sufficient to accomplish knowledge transfer from the training scenes to previously unseen test scenes. We believe that this general idea might be applicable beyond head portrait reconstruction to other classes (e.g.~full-body reconstruction) and architectures (e.g.~different NeRF types).

\begin{figure}[!h]
    \centering
    \includegraphics[trim={0 100 170 15},clip,width=0.16\textwidth]{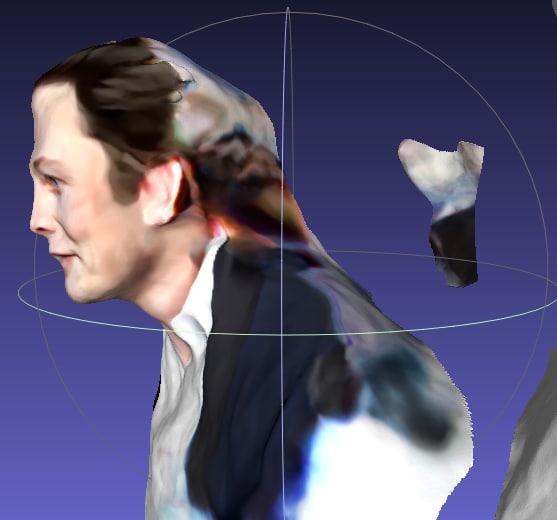}
    \caption{A limitation of training on SmartPortraits: the back is never visible in the dataset, leading to poor reconstructions or occluded regions beyond the ears.}
    \label{fig:Failure_cases_back}
\end{figure}

Our approach has certain limitations. Many of them are due to rather constrained training dataset. First, there are only 103 training sequences. Although Multi-NeuS' generalization ability seems very good for such a small dataset, there is still low diversity of hair styles, adornments, and skin types. In addition, SmartPortraits only exhibits neutral facial expressions, though in practice Multi-NeuS still seems to reconstruct smiles reasonably well. Moreover, the camera in the dataset only travels at most $\pm 45^{\circ}$ around the head and therefore does not capture the back. As a result, our model always fails to reconstruct the back because it has never "seen" it in training (\fig{In-the-wild reconstruction}, bottom right; \fig{Failure_cases_back}). Thus, an obvious remedy to improve the quality is to expand our training set.

\begin{figure}[!h]
    \centering
    \includegraphics[trim={0 0 0 0},clip,width=0.47\textwidth]{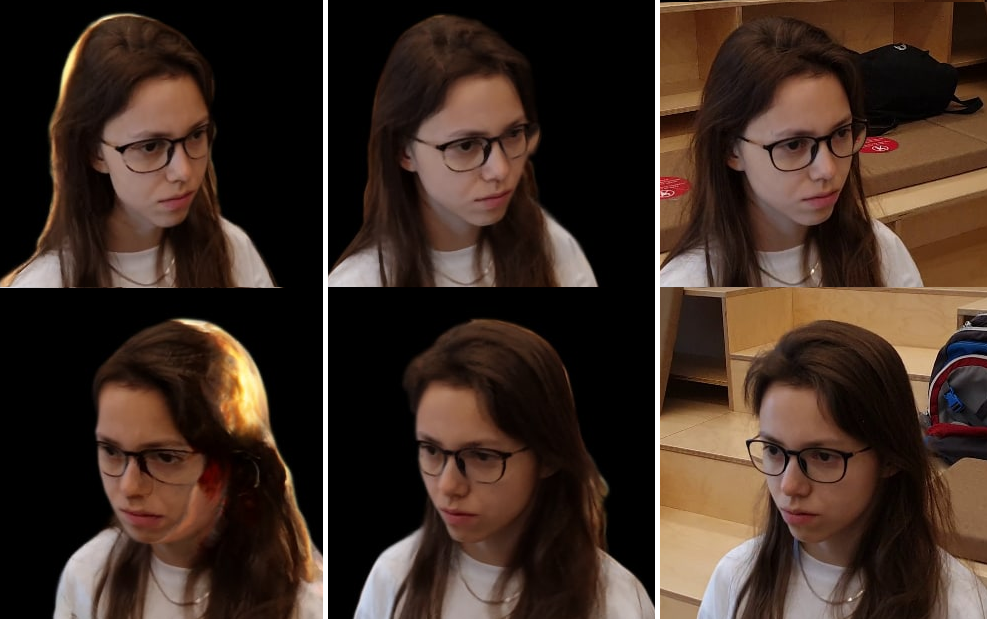}
    \caption{A limitation of the underlying architecture of Multi-NeuS: at the second step of fitting, artifacts sometimes reappear in the occluded regions (left column: left cheek and "yellow hair"). Shown are the renders of the control views (top and bottom) for Multi-NeuS fitted to one frontal (left) and two $\pm 45^{\circ}$ (middle) views next to the ground truth (right). Here we apply our best model on a validation scene from SmartPortraits.}
    \label{fig:Failure_cases_1_vs_2_views}
\end{figure}

Our models might benefit greatly from further improvements and simplifications of the underlying architecture. While the first step of fitting often provides a good initialization for occluded regions, the second step sometimes worsens these regions (\fig{SmartPortraits}; \fig{Failure_cases_1_vs_2_views}). This could be addressed with ad-hoc inpainting procedures that exploit class-specific symmetries, or more principled extensions of our method such as learned gradient descent \cite{Andrychowicz16, Flynn19}. However, the fundamental problem might be hidden deeper in the network architecture. This is additionally highlighted by the fact that Multi-NeuS struggles to fit training samples with the same accuracy as NeuS. A promicing direction for future investigation is therefore how to reduce the model complexity even further (e.g.\ by using small learnable latent dictionaries) and to allow for very large datasets and better generalization.

Finally, the models produced with our approach come without rigging capability, and in the future it would be interesting to extend our framework to address this.


\section{Acknowledgment} The Authors acknowledge the use of computational resources of the Skoltech supercomputer Zhores \cite{zacharov2019zhores} for obtaining the results presented in this paper.

%% file: sec/6_authors.tex
\begin{IEEEbiography}[{\includegraphics[width=1in,height=1.25in,clip,keepaspectratio]{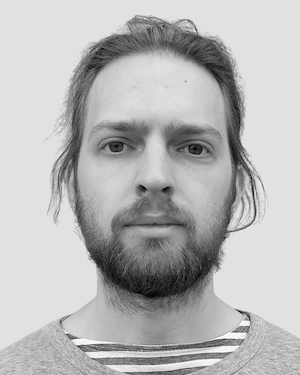}}]{Egor Burkov
} received the B.S. degree in applied mathematics and information science from the Higher School of Economics, Moscow, Russia in 2016. Since then, he has been with the Skolkovo Institute of Science and Technology, Moscow, Russia, receiving the M.S. degree in data science in 2018 and currently pursuing a Ph.D. degree in computational and data science and engineering. His research interests include computer vision, especially 3D vision and neural rendering, as well as general deep learning, notably generative models and computationally efficient neural architectures.
\end{IEEEbiography}

\begin{IEEEbiography}[{\includegraphics[width=1in,height=1.25in,clip,keepaspectratio]{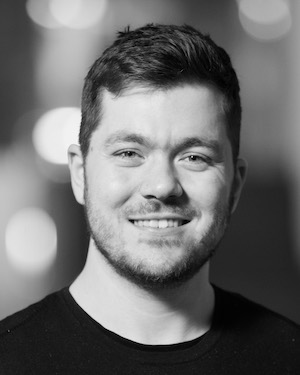}}]{Ruslan Rakhimov}  received the B.Sc.
degree in applied physics and mathematics from the
Moscow Institute of Physics and Technology, in 2018, and the M.Sc. degree in data
science from the Skolkovo Institute of Science and
Technology, in 2020, where he
is currently pursuing the Ph.D. degree in computational and data science and engineering.
His main research interests include 3d computer vision and deep learning.
\end{IEEEbiography}

\begin{IEEEbiography}[{\includegraphics[width=1in,height=1.25in,clip,keepaspectratio]{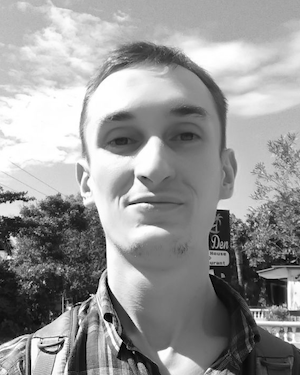}}]{Aleksandr Safin} is a doctoral student at Skoltech, working on the junction of unpaired learning and methods for 3d reconstruction. Prior to that, he collaborated with Huawei Research Center on computational photography approaches. Aleksandr obtained his master’s from Skoltech in 2019, and bachelor’s from Higher School of Economics in 2017.
\end{IEEEbiography}

\begin{IEEEbiography}[{\includegraphics[width=1in,height=1.25in,clip,keepaspectratio]{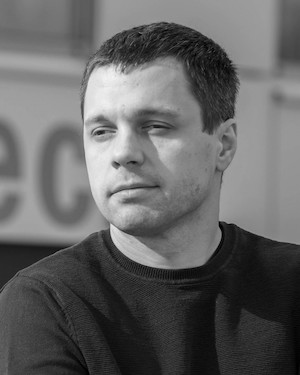}}]{Evgeny Burnaev}  received the M.Sc. degree
in applied physics and mathematics from the
Moscow Institute of Physics and Technology,
in 2006, and the Ph.D. degree in foundations of
computer science from the Institute for Information Transmission Problem RAS, in 2008.
He is currently works at the Skolkovo Institute of Science and Technology
as a Full Professor and a Director of Skoltech Applied AI center. His research interests
include regression based on Gaussian processes and kernel methods for
multi-fidelity surrogate modeling and optimization, deep learning for 3D
Data analysis and manifold learning, on-line sequence learning for prediction, and non-parametric anomaly detection.
\end{IEEEbiography}

\begin{IEEEbiography}[{\includegraphics[width=1in,height=1.25in,clip,keepaspectratio]{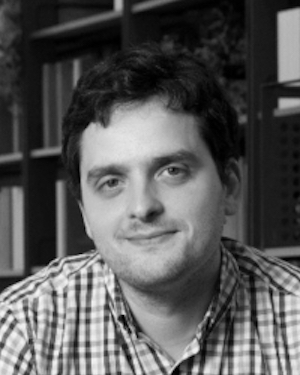}}]{Victor Lempitsky} received the Ph.D. degree from Moscow State University, in 2007. He was an associate professor and the head of the Computer Vision group, Skolkovo Institute of Science and Technology (Skoltech). He also was a lab leader at Samsung AI Center, Moscow and later a lab leader at Yandex. His research interests include computer vision, biomedical image analysis, and deep learning.
\end{IEEEbiography}